\documentclass[letterpaper]{article} 
\usepackage{a2026}  
\usepackage{times}  
\usepackage{helvet}  
\usepackage{courier}  
\usepackage[hyphens]{url}  
\usepackage{graphicx} 
\usepackage{natbib}  
\usepackage{caption} 
\usepackage{algorithm}
\usepackage{algorithmic}

\usepackage{newfloat}
\usepackage{listings}
\usepackage{bm}
\usepackage{amssymb}
\usepackage{amsmath}
\usepackage{booktabs}
\usepackage{caption}
\usepackage[table]{xcolor}
\DeclareCaptionStyle{ruled}{labelfont=normalfont,labelsep=colon,strut=off} 
\floatstyle{ruled}
\newfloat{listing}{tb}{lst}{}
\floatname{listing}{Listing}

\newcommand{\refrn}{{\bm{\hat{\theta}}_{t+1}}}
\newcommand{\simr}{{\bm{{\theta}}_{t}}}
\newcommand{\reftn}{{\bm{\hat{p}}_{t+1}}}
\newcommand{\simt}{{\bm{{p}}_{t}}}
\newcommand{\simp}{{\bm{{q}}_{t}}}
\newcommand{\state}{{\bm{s}_t}}
\newcommand{\selfstate}{{\bm{s}^{\text{p}}_t}}
\newcommand{\objectstate}{{\bm{s}^{\text{g-obj}}_t}}
\newcommand{\goalstate}{{\bm{s}^{\text{g}}_t}}
\newcommand{\simv}{{\bm{\dot{q}}_{t}}}

\newcommand{\objrefrn}{{\bm{\hat{\theta}}^\text{obj}_{t+1}}}
\newcommand{\objsimr}{{\bm{{\theta}}^\text{obj}_{t}}}
\newcommand{\objreftn}{{\bm{\hat{p}}^\text{obj}_{t+1}}}
\newcommand{\objsimt}{{\bm{{p}}^\text{obj}_{t}}}

\nocopyright 

\title{
SimGenHOI: Physically Realistic Whole-Body Humanoid-Object Interaction \\ via Generative Modeling and Reinforcement Learning
}
\author{
    Yuhang Lin\textsuperscript{\rm 1},
    Yijia Xie\textsuperscript{\rm 1},
    Jiahong Xie\textsuperscript{\rm 1},
    Yuehao Huang\textsuperscript{\rm 1},\\
    Ruoyu Wang\textsuperscript{\rm 1},
    Jiajun Lv\textsuperscript{\rm 1},
    Yukai Ma\textsuperscript{\rm 1},
    Xingxing Zuo\textsuperscript{\rm 2}\thanks{Corresponding author}
}
\affiliations{
    \textsuperscript{\rm 1}Zhejiang University\\
    \textsuperscript{\rm 2}Mohamed bin Zayed University of Artificial Intelligence (MBZUAI)\\

}

\usepackage{bibentry}

\begin{document}

\maketitle

\begin{abstract}
 Generating physically realistic humanoid–object interactions (HOI) is a fundamental challenge in robotics. Existing HOI generation approaches, such as diffusion-based models, often suffer from artifacts such as implausible contacts, penetrations, and unrealistic whole-body actions, which hinder successful execution in physical environments. To address these challenges, we introduce \textbf{SimGenHOI}, a unified framework that combines the strengths of generative modeling and reinforcement learning to produce controllable and physically plausible HOI. Our HOI generative model, based on Diffusion Transformers (DiT), predicts a set of key actions conditioned on text prompts, object geometry, sparse object waypoints, and the initial humanoid pose. These key actions capture essential interaction dynamics and are interpolated into smooth motion trajectories, naturally supporting long-horizon generation. To ensure physical realism, we design a contact-aware whole-body control policy trained with reinforcement learning, which tracks the generated motions while correcting artifacts such as penetration and foot sliding. Furthermore, we introduce a mutual fine-tuning strategy, where the generative model and the control policy iteratively refine each other, improving both motion realism and tracking robustness. Extensive experiments demonstrate that SimGenHOI generates realistic, diverse, and physically plausible humanoid–object interactions, achieving significantly higher tracking success rates in simulation and enabling long-horizon manipulation tasks. Code will be released upon acceptance on our project page: {\ttfamily https://xingxingzuo.github.io/simgen\_hoi/}.
\end{abstract}

\section{Introduction}
    Generating humanoids-object interaction (HOI) sequences with specific conditionals is a pivotal research domain in computer vision, embodied AI, and robotics. While recent studies have made progress in generating HOI sequences~\citep{cen2024generating,yi2024generating}, they often focus on simple actions in static scenarios (e.g., lying on a bed), and overlook the interaction with dynamic objects in everyday real-world settings.
    
    \begin{figure}[htp]
    \centering
    \vspace{-5mm}
    \includegraphics[width=\linewidth]{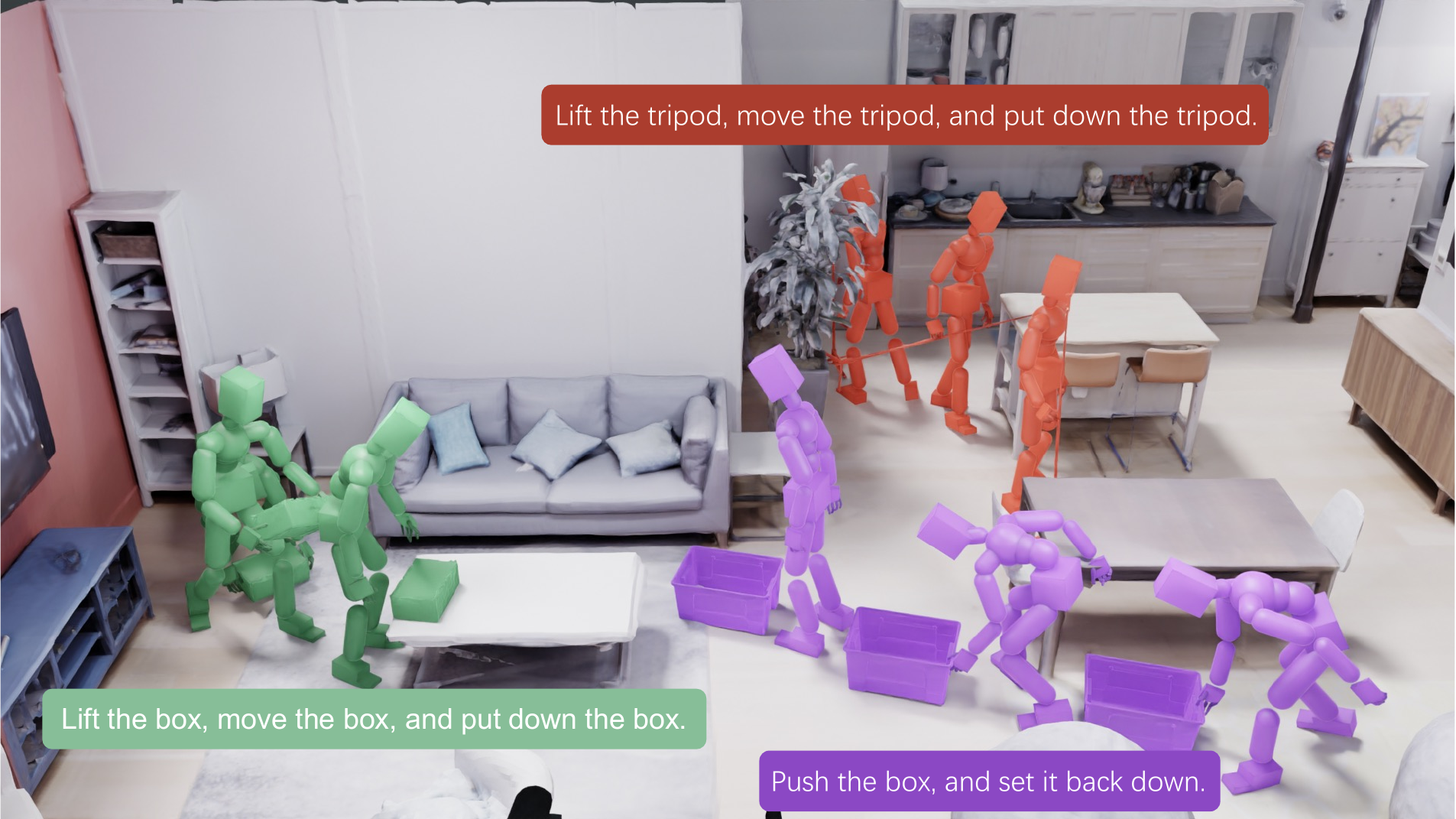}
    \caption{With the condition of text prompt,  object geometry, sparse waypoints of the object, and initial humanoid pose, SimGenHOI can generate the whole-body humanoid-object interaction sequence and employ a contact-aware control policy to track the generated trajectory in a physically realistic way.} 
    \label{fig:main}
    \end{figure}
    
    The field currently faces a fundamental trade-off between generative diversity and physical realism. On the one hand, generative models excel at producing varied and controllable motions. Some approaches utilize egocentric 3D occupancy information and multiple action labels~\citep{truman2024,lingo2024}, the movement of the objects is optimized based on the hands' motion, while others employ diffusion models to co-synthesize human actions and object trajectories conditioned on sparse waypoints~\citep{chois2024}. Subsequent work also integrates state-of-the-art grasping generators~\citep{hoifhli2024} following a coarse-to-fine pipeline. However,  methods based on generative models struggle to eliminate physical artifacts like foot sliding or hand penetration. On the other hand, reinforcement learning (RL) can simulate physically plausible interactions governed by physical rules. Closed~\citep{Closd2024} effectively integrates a motion diffusion model with a tracking policy~\cite{Luo2023PerpetualHC}, which results in a more robust and physically consistent framework for humanoid motion generation. However, it is limited to simple interactions, such as hitting something, and fails to generalize to highly dynamic, long-horizon tasks.
    
    This work aims to bridge this gap by proposing a framework that combines the strengths of generative models with physics-based pose policies learned through reinforcement learning. Our goal is to generate realistic, long-horizon, and dynamic HOI sequences that are not only controllable but also physically realistic and executable in a physics-aware simulator.

    To achieve this goal, we employ a Diffusion Transformers model~\cite{DiT} conditioned on a text prompt, object geometry, and sparse waypoints of the object to generate a sequence of human motions, object motions and contact probability. 
    To enforce physical plausibility of the generated sequence, we incorporate a contact-aware humanoid whole-body controller that corrects artifacts like penetration and foot sliding in the generated motions. Finally, to mitigate tracking failures from Out-of-distribution(OOD) states, we propose a mutual fine-tuning strategy that iteratively refines both the generation model and the contact-aware policy.
    The main contributions of this work can be summarized as follows.
\begin{itemize}
    \item We propose a physically realistic whole-body action generative framework for humanoid-object interactions by articulating a diffusion based HOI generative model and a contact-aware whole-body control policy. Specifically, instead of generating dense motion sequences, our HOI generative model predicts a set of key actions that capture the essential interaction dynamics. These key actions are then interpolated to produce complete and smooth motion trajectories for the HOI policy to track. This paradigm naturally supports long-horizon HOI generation and significantly enhances physical feasibility.
    \item We design a contact-aware whole-body control policy for complex humanoid-object interaction that supports a broader spectrum of actions beyond simple object transportation.
    \item We further fine-tune the generative model using the actions successfully tracked by our contact-aware HOI policy. Conversely, the more realistic actions from the improved generative model can further enhance the HOI policy, leading to a higher tracking success rate.
    \item We validate the effectiveness of our proposed framework through extensive experiments, demonstrating its ability to generate realistic, diverse, and physically plausible humanoid-object interactions across various scenarios.
\end{itemize}

\begin{figure*}
    \centering
    \vspace{-7mm}
    \includegraphics[width=0.95\linewidth]{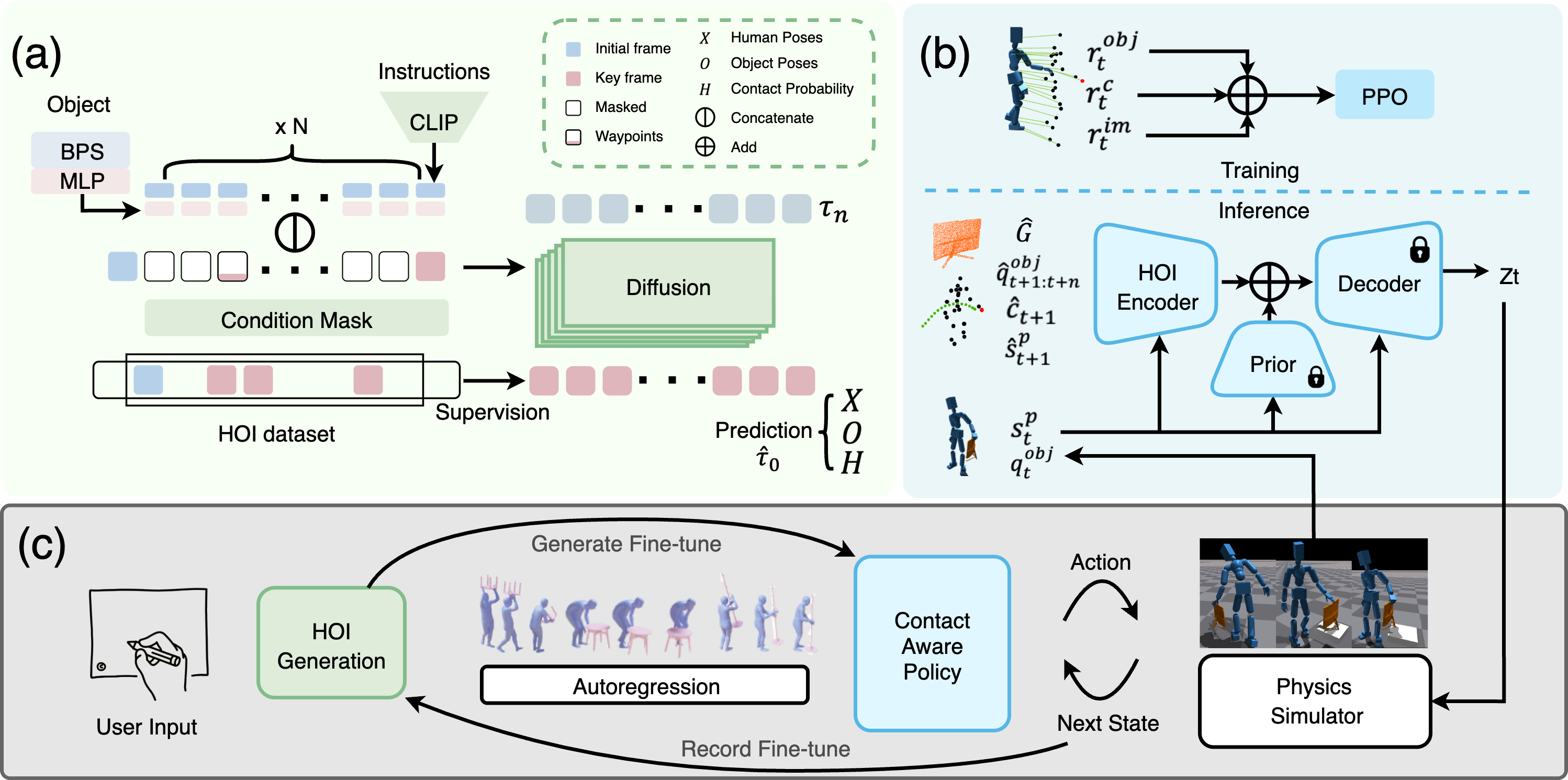}
    \caption{\textbf{Our proposed framework uses a diffusion model for key action generation and reinforcement learning to train a contact-aware HOI policy in the Isaac Gym simulator.} The proposed framework, as depicted in (c), mainly consists of two components, including: \textbf{(a)} HOI generation conditioned on text prompt, object geometry, sparse waypoints of objects, and initial humanoid pose. \textbf{(b)} Contact-aware humanoid-object interaction (HOI) policy; $\bm{\hat G}$, $\bm {\hat q}^{obj}_{t+1:t+n}$ and $\bm q_t^{obj}$ represent the object's geometry, future trajectory and its current state. $\bm {\hat{c}}_{t+1}$, $\bm s^p_t$ and $\bm {\hat s}_{t+1}^p$ denote the contact guidance, current state of the humanoid, and its next state reference.}
    \label{fig:pipline}
\end{figure*}
\section{Related Works}
    \noindent \textbf{Data-driven motion synthesis.}
    In recent years, significant advances have been made in data-driven motion
    synthesis. The availability of large-scale motion capture datasets, such as AMASS~\citep{AMASS}, and motion-language annotation datasets, like BABEL~\citep{BABEL} and HumanML3D~\citep{Guo_2022_CVPR}, has catalyzed rapid progress in text-conditioned motion generation. Early methods predominantly relied on Variational Autoencoders (VAEs) to map natural language descriptions to latent motion representations. More recently, diffusion models have emerged as the state-of-the-art approach~\citep{zhang2024motiondiffuse,dabral2023mofusion}, offering substantial improvements in the realism and diversity of generated motions. Seminal works in this area include MDM by Tevet et al.~\citep{tevet2022human}, which generates high-quality motion sequences, and the work of Raab et al. on MoMo~\citep{raab2024monkey} and SinMDM~\citep{raab2023single}, which demonstrate strong generalization to out-of-distribution textual inputs. Furthermore, A-MDM~\citep{shi2024interactive} and CAMDM~\citep{chen2024taming}, proposed by Shi and Chen, respectively, leverage autoregressive diffusion to enable interaction-aware motion control. Collectively, these developments provide a robust foundation for downstream applications such as motion editing, interaction modeling, and trajectory planning.

    \noindent \textbf{Diffusion models for object interaction.}
    Diffusion-based object interaction modeling can be broadly categorized into human-object interaction (HOI) and human-scene interaction (HSI). For static interaction scenarios, such as generating poses for sitting on chairs or lying on sofas, conditional diffusion models have demonstrated strong performance. Works by Wu~\citep{wu2024thor}, Peng~\citep{peng2023hoi}, among others, have achieved high-quality pose synthesis and realistic pose synthesis. In more dynamic settings, recent approaches like OMOMO~\citep{li2023object} predict full-body motion conditioned on object trajectories. Other methods focus on different conditioning signals. For instance, InterDiff~\citep{xu2023interdiff} focuses on predicting interaction from motion history, while CHOIS~\citep{li2024controllable} incorporates language and waypoint-based cues for human-object interaction synthesis. Meanwhile, methods that leverage physical priors, such as the work of Huang et al.~\citep{huang2023diffusion}, offer improved realism but often struggle to balance physical plausibility with multi-modal controllability.
    
    \noindent \textbf{Physics-Based Humanoid-Object Interactions.}
    Physics-based interaction is a central topic in humanoid control, encompassing tasks from dexterous manipulation and whole-body control. Deep reinforcement learning (DRL) has been widely adopted for learning whole-body control policies due to its flexibility in handling complex dynamics~\citep{chentanez2018physics, peng2018deepmimic}. However, fine-grained control, particularly of the hands and fingers, remains a significant challenge~\citep{bae2023pmp, braun2024physically, liu2018learning, merel2020catch}. In dexterous manipulation, proposed methods range from kinematic models~\citep{romero2022embodied} to physics-based simulations~\citep{liu2018learning, wang2023physhoi}. Despite progress, achieving policy generalization and scalability remains a primary obstacle~\citep{merel2020catch}. Recent advancements include full-body trajectory tracking via latent space control~\citep{luo2023universal} and hybrid frameworks that combine planning and optimization~\citep{xu2023interdiff}. Nevertheless, accurately simulating dynamic interactions~\citep{kulkarni2024nifty,zhang2022couch} and bridging the sim-to-real gap~\citep{he2024omnih2o} are persistent open problems, motivating the need for tighter integration of physics and kinematic priors.
\section{Preliminaries}
\label{sec:preliminaries}
    \noindent \textbf{Motion Representation.}
    We use two different representations of motion, each suitable for its purpose. The HOI generative model produces both the human and object poses. The human poses are represented by $\bm X\in \mathbb{R}^{T\times D}$, where $ T$ and $ D$ represent the time steps and dimension of the human pose. $\bm{X}_t$, corresponding to the human pose at time $t$, consists of joint positions and 6 degree-of-freedom (DOF) rotations~\cite{zhou2019continuity} compatible with the SMPL-X model~\cite{smplx}. Object poses are represented by the global position and the rotation matrix relative to the input object geometry as $\bm O\in \mathbb{R}^{T\times 12}$. 
    
    For training the HOI policy in physical simulation, we represent the humanoid motion $\bm x^{\text{sim}}$ that consists of joints rotation $\bm \theta_t \in \mathbb{R}^{J\times 6}$ using the 6 DOF rotation representation~\citep{zhou2019continuity}, position $\bm p_t \in \mathbb{R}^{J\times 3}$, linear velocity $\bm v_t \in \mathbb{R}^{J\times 6}$, and angular velocity $\bm \omega_t \in \mathbb{R}^{J\times 6}$ of all $J$ links in the humanoid (hands and body). The object poses are represented by $\bm q^\text{obj}_t$, comprised of global position $\bm p^\text{obj}_t$, orientation $\bm \theta^\text{obj}_t$, linear velocity $\bm v^\text{obj}_t$, and angular velocity $\bm \omega ^\text{obj}_t$. We convert the human pose generated by the diffusion model using $\bm{x}^\text{sim} = R2G(x)$, a relative-to-global transformation, with its inverse defined as $\bm{x} = G2R(\bm{x}^\text{sim})$.

    \noindent \textbf{Object Geometry Representation.}
    We leverage the Basis Point Set (BPS)~\citep{bps} to represent the geometric properties of the object. Following prior work OMOMO~\citep{li2023object}, we sample a set of basis points from the volume of a sphere with a 1-meter radius and calculate the minimum Euclidean distance to the nearest point on the object's mesh. Then, the directional vectors from the basis points to their nearest neighbors are recorded as $\bm{G} \in \mathbb{R}^{1024\times 3}$. Finally, an MLP projects the vectors to a low-dimensional object geometry as $\bm{\hat G} \in \mathbb{R}^{256\times 3}$.
\section{Methods}
This section details our method for generating physically plausible humanoid-object interactions. The overall framework is illustrated in Figure~\ref{fig:pipline}. First, we build an HOI generation model to generate a sequence of humanoid actions and corresponding object trajectories based on given conditions. Then we incorporate contact guidance to train a whole-body control policy to track the generated HOI seqequence. Notably, we introduce a mutual fine-tuning process between the generative model and the control policy, leading to improved performance through iterative refinement. The pseudocode of our proposed method is listed in alg~\ref{alg:sim_gen_hoi}.

\subsection{Humanoid-Object Interaction Generation.}
    \noindent \textbf{Interaction Diffusion Model.}
    We adopt a conditional diffusion model based on the DiT architecture~\citep{DiT}, which uses a Transformer~\cite{transformer} as the denoising network to jointly generate human motion and object trajectories. Our framework is built upon Denoising Diffusion Probabilistic Models (DDPM)~\citep{ddpm}, which model generation as the reversal of a fixed  Markovian noising process. In this process, a clean data sample $\bm \tau_{0}$ is gradually corrupted with Gaussian noise over $N$ steps:
    \begin{align}
    \label{eq:forward_diffusion_step}
        q(\bm{\tau}_{n}|\bm{\tau}_{n-1}) := \mathcal{N}(\bm{\tau}_{n}; \sqrt{1-\beta_{n}}\bm{\tau}_{n-1}, \beta_{n}\bm{I}), \\
         q(\bm{\tau}_{1:N}|\bm{\tau}_{0}) := \prod_{n=1}^{N} q(\bm{\tau}_{n}|\bm{\tau}_{n-1}),
    \end{align}
    where $\beta_{n}$ represents a fixed variance schedule and $\bm{I}$ is an identity matrix. Our goal is to learn a model $p_{\theta}$ with param $\theta$ to reverse the forward diffusion process, conditioned on a set of signals $\bm C$:
    \begin{equation}
    \label{eq:reverse_diffusion_step}
        p_{\theta}(\bm{\tau}_{n-1}|\bm{\tau}_{n}, \bm{C}) := \mathcal{N}(\bm{\tau}_{n-1}; \bm{\mu}_{\theta}(\bm{\tau}_n, n, \bm{C}), \bm{\Sigma}_{n}),
    \end{equation}
    where $\bm{\mu}_{\theta}$ denotes the predicted mean and $\bm{\Sigma}_{n}$ is a fixed variance. Following prior work~\citep{ddpm}, we re-parameterize the model to predict the original clean data ${\bm{\tau}_0}$ from the noisy input $\bm{\tau}_n$. The model is trained with the objective:
    \begin{equation}
    \label{eq:loss}
         \mathcal{L} = \mathbb{E}_{\bm{\tau}_0, n}||\hat{\bm{\tau}}_{\theta}(\bm{x}_{n}, n, \bm{C}) - \bm{\tau}_{0}||_{1}.
    \end{equation}
    where $\hat{\bm\tau_\theta}$ represents the final prediction of the model.
    
    The conditions $\bm{C}$ include the object's BPS geometry embedding ${\bm{\hat G}}$, masked motions $\bm{S}$ for in-filling or editing tasks, and a text prompt embedding $\bm {e}_{\text{text}}$ from a pre-trained CLIP~\citep{clip} model.  $\bm{S}\in \mathbb{R}^{T \times (12+D)}$ is the masked poses of human and object,  while preserving their initial poses, sparse waypoints of 2D object positions(x,y), and the target 3D object position(x,y,z). The noise level $n$ is embedded via an MLP and fused with the language embedding to form a unified conditioning vector, which is passed to the Transformer network alongside the sequence.  The diffusion model predicts  the human poses $\bm{X}$, object poses $\bm O$ and also the contact probability of hands and feet, which is denoted as $\bm H \in \mathbb{R}^{T\times 4}$. The model architecture is shown in Figure~\ref{fig:pipline}(a).

    \begin{figure}[t]
    \centering
    \vspace{-7 mm}
    \setlength{\abovecaptionskip}{5pt}
    \includegraphics[width=\linewidth]{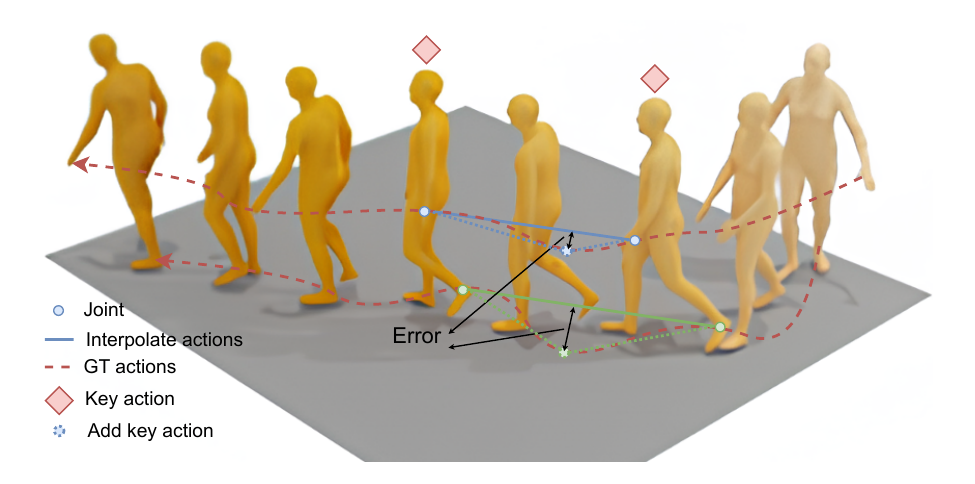}
    \caption{\textbf{Reconstruction Error} of humanoid actions for key action extraction.} 
    \label{fig:keyframe_extract}
    \end{figure}

    \noindent \textbf{Key Actions Extract.}
    Inspired by AWE~\citep{shi2023waypoint}, which predicts un-uniform actions from the current robot state, we adapt this concept to extract a sparse set of key actions from dense humanoid action sequences. This reformulates the generation problem from predicting dense uniform motions to predicting a sequence of critical actions. Special attention is paid to accurately recovering fine-grained details of critical joints such as the hands and feet. 

    Given a dense motion sequence for a set of joints, $\bm{J}=\{\bm j_t\}_{t=0}^{|\bm J|-1}$ where $\bm j_t \in \mathbb{R}^{N\times 3}$, we aim to find a sparse sequence of key actions $\bm{K}=\{k_i\}^{|\bm K|-1}_{i=0}$. By linearly interpolating these key actions, we can reconstruct an approximate motion sequence $\bm{\hat J} = f(\bm K) = \{\bm {\hat j}_t\}_{t=0}^{|\bm J|-1}$. The whole sequence reconstruction error is the maximum of each subsequence split by key actions $\bm K$. The quality of the key actions in each subsequence is measured by a weighted reconstruction error, with higher weights \( \bm w \in \mathbb{R}^{N} \) assigned to our selected critical joints:
    \begin{align}
    \label{eq:reconstruction_error}
    \mathcal{L}(\hat{\bm J}, \bm J) = \max_{t=0,...,|\bm J|-1} (\max_{n\in \{1,...,N\}} ({w}^n \ell(\bm j^n_t, \hat{\bm j}^n_t)))
    \end{align} 
    where $\bm J$ is the set of humanoid joints, \( \bm j_t \) and \( \bm {\hat{j}_t} \) denote the ground-truth and interpolated trajectories of joints, and \( \ell(\cdot, \cdot) \) computes the Euclidean distance. \( \bm w \) is a manually defined importance weight,  where $w^n$ represents its $n$-th element. The diagram to illustrate reconstruction error is shown in Figure~\ref{fig:keyframe_extract}.
    
    With the reconstruction error in Eq.\eqref{eq:reconstruction_error} available, we can automatically extract key actions from an entire motion sequence using a simple recursive algorithm. The first and last frames are always treated as initial key actions for any given motion sequence, forming the first interpolated actions. The frame with the largest reconstruction error along the actions is then identified and added as a new key action, effectively dividing the sequence into two sub-segments. The same procedure is recursively applied to each sub-segment: compute the interpolated actions, identify the frame with the highest error, and insert it as a key action. This process continues until the reconstruction error for all sub-segments falls below a predefined threshold, at which point key actions extraction is considered complete.
    
    \noindent \textbf{Training.}
    Following key action extraction, we curate the training dataset using a sliding window applied to the original action sequences. In each window, the first frame represents the initial state, and the remaining frames are populated with the extracted key actions. Every sliding window serves as a data sample for the training of our generation model. This structure enables us to train the model for predicting a sequence of upcoming key actions. The model is trained using the L1 loss between the predictions and the ground-truth.

    

\subsection{Contact-Aware Whole Body Policy.}
    \label{sec:hoi_policy}
    To execute the generated HOI sequence in a physically realistic environment, we develop a contact-aware whole-body control policy. Our policy leverages the pre-trained humanoid motion representation model PULSE~\citep{luo2023universal}, trained on the large-scale AMASS dataset~\citep{AMASS}, which offers strong motion priors for downstream humanoid-object interaction tasks. The detailed structure is illustrated in Figure~\ref{fig:pipline}(b).

    \noindent \textbf{Human Motion Tracking.}
    In a physically realistic simulator, we can get the humanoid state $\state \triangleq (\selfstate, \goalstate)$ comprised of humanoid proprioception $\selfstate$ and the reference humanoid state $\goalstate$ at time $t$. The proprioceptive state $\selfstate \triangleq (\simp, \simv, \bm \beta)$ contains the position of the humanoid's body links $\simp$, velocity $\simv$, along with optional body shapes $\bm \beta$. 
    $\bm \beta$ contains information about the length of the limb of each body link \cite{luo2022universal} to accommodate different body shapes during the training.
    The reference state $\goalstate$ is defined as:
    \begin{equation}
        \begin{aligned}
        \bm{s}^\text{g}_t \triangleq & (  \refrn \ominus \simr, \reftn -  \simt, \bm{\hat{v}}_{t+1} -  \bm{v}_t,  \\ & \bm{\hat{\omega}}_{t+1} -  \bm{\omega}_t, \refrn, \reftn), 
        \end{aligned}
    \end{equation}
    where $\refrn, \reftn, \bm{\hat{v}}_{t+1}, \bm{\hat{\omega}}_{t+1}$ are the orientation, position, linear velocity, and angular velocity for the next reference motion. $\ominus$ calculates the rotation difference. The reference velocities are calculated from kinematic rules based on the given reference trajectory.
    
\begin{figure}[t]
    \centering
    \vspace{-7 mm}    
    \includegraphics[width=0.45\textwidth]{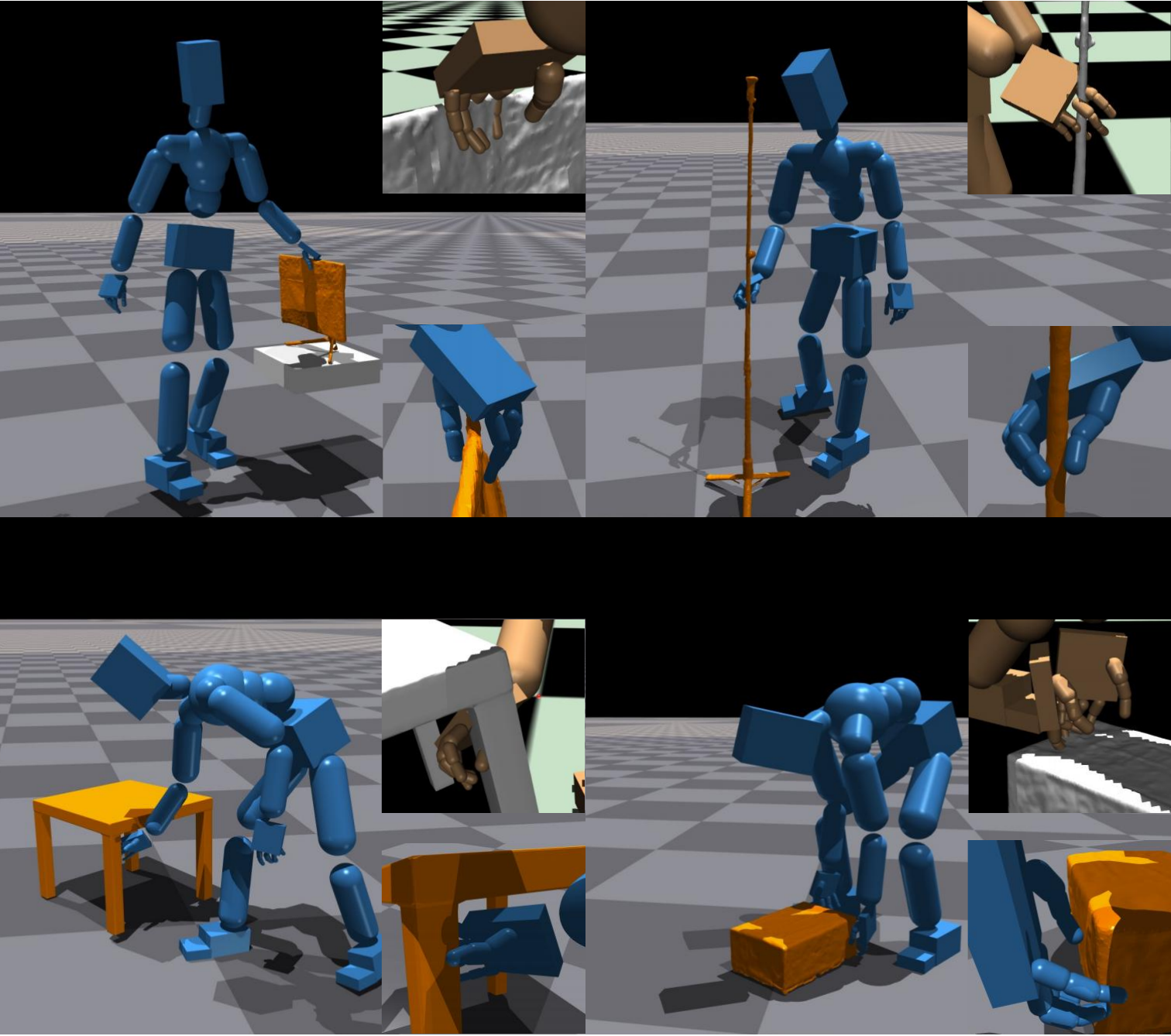}
    \caption{\textbf{Qualitative results} of combining with simulation. The detailed hand-object interaction is zoomed in the bottom-right corner, while the top-right corner visualizes the OMOMO dataset~\cite{li2023object} grasping motion refined by GRIP~\cite{taheri2024grip}.} 
    \label{fig:qualitative_res}
\end{figure}

    \noindent \textbf{Contact Guidance Manipulation.}
    Building upon motion tracking, we incorporate object observations by constructing representations of object state in addition to the contact guidance $\bm{c}_t$ between the humanoid and the objects. This contact guidance encodes latent actions for the humanoid, such as grasping or releasing, and provides temporal guidance for when interactions should occur. This design not only enriches the representation of human-object interactions but also bridges the generative model and the HOI policy by providing consistent guidance signals. The simulation state  $\bm s_t^\text{obj}$ of the object consists of the object's current state $\bm s_t^\text{p-obj}$ and reference state  $\bm s_t^\text{g-obj}$. The object current state $\bm s_t^\text{p-obj}$ contains the position of object $\objsimt$, velocity $\bm{v}^\text{obj}_t$ and the Boolean contact guidance $\bm{c}_t$. With $\odot$ represents XNOR, the complete object observation is defined as
    \begin{equation}
        \begin{aligned}
        \objectstate \triangleq & (  \objrefrn \ominus \objsimr, \objreftn -  \objsimt, \bm{\hat{v}}^\text{obj}_{t+1} -  \bm{v}^\text{obj}_t,  \\ & \bm{\hat{\omega}}^\text{obj}_{t+1} -  \bm{\omega}^\text{obj}_t, \bm{\hat c}_{t+1} \odot \bm{c}_{t}, \objrefrn, \objreftn, \bm{\hat c}_{t+1}),
        \end{aligned}
    \end{equation}

    \noindent \textbf{Reward Design.}  
    Since the generative model does not predict reference for detailed hand poses, our policy only tracks the predicted poses of key joints. Please see the appendix E for details of the key joints. However, the lack of grasping details makes accurate object manipulation challenging. To address this issue, we design a reward function to encourage the humanoid to make contact with objects. For tasks that do not require high dexterity, contact-promoting rewards can help the policy develop effective hand interaction strategies, thereby fully leveraging the exploratory nature of reinforcement learning. Specifically, the tracking reward for tracking human motion is defined as:

    \begin{equation}
        \begin{aligned}
         \bm r^{\text{g}}_t & = \bm w_{\text{jp}} e^{-100 \| \bm{\hat p}_t - \bm{p}_t \|}  + \bm  w_{\text{jr}} e^{-10\| \bm{\hat q}_t \ominus \bm{q}_t \|} \\ &  + \bm  w_{\text{jv}} e^{-0.1\| \bm{\hat v}_t - \bm{v}_t \|} + \bm  w_{\text{j}\omega} e^{-0.1\| \bm{\hat \omega}_t - \bm{\omega}_t \|} \\ & + \bm  w_{\text{jc}} \|\bm {\hat c}_t \odot \bm c_t\|
        \end{aligned}
    \end{equation}
    This reward is only applied to some of our selected key joints. This encourages the humanoid to accurately track the reference humanoid states, and correct contact of bodies over time. 
    
    The object-related reward $\bm{r}^{\text{obj}}_t$ is similar to tracking reward, which encourages the same object positions, orientations, linear and angular velocities as reference motions.
    \begin{equation}
        \begin{aligned}
         \bm r^{\text{g-obj}}_t & = \bm w_{\text{op}} e^{-100 \| \bm{\hat {p}}^{obj}_t - \bm{p}^{obj}_t \|}  + \bm w_{\text{or}} e^{-10\| \bm{\hat {q}}^{obj}_t \ominus \bm{q}^{obj}_t \|} \\ &  + \bm w_{\text{ov}} e^{-5\| \bm{\hat {v}}^{obj}_t - \bm{v}^{obj}_t \|} + \bm w_{\text{o}\omega} e^{-5\| \bm{\hat {\omega}}^{obj}_t - \bm{\omega}^{obj}_t \|}
        \end{aligned}
    \end{equation}
    
    The overall reward used to train the policy is weighted as follows:
    \begin{equation}
    \bm{r}_t = \bm \alpha \bm{r}^{\text{g}}_t + (1-\bm \alpha) \bm{r}^{\text{g-obj}}_t
    \end{equation}
    
    \noindent \textbf{Early Termination.}
    Early Termination (ET)~\citep{peng2018deepmimic} is a standard technique in motion tracking, where an episode is terminated early if the character makes unintended ground contact or strays too far from the reference motion~\citep{Luo2023PerpetualHC}. This prevents the policy from learning from unrealistic or undesirable states. However, in scenarios involving humanoid-object interaction, additional termination criteria are necessary. To address this, we introduce Interaction Early Termination, which extends ET by triggering termination under two additional conditions: (i) the average deviation of object keypoints from their reference positions exceeds 0.5 meters; (ii) expected body-object contact is absent for over 10 consecutive frames.

\subsection{HOI In Physics Simulation.}
\label{section:finetune}
    \noindent \textbf{Fine-Tuning.}
    To synergize the generative model and the tracking policy, we evaluate the policy within a physics-aware simulator to track the generated motions and collect physically plausible actions that adhere to contact constraints. These executable and physically-plausible actions serve as high-quality samples for fine-tuning the diffusion-based generative model, enhancing its ability to generate more physically feasible and easier to track motions. Conversely, the tracking policy is further fine-tuned with the motion generated from the improved generative model. This alternating fine-tuning process helps align the distribution of the generated motion and the policy behavior, improving consistency between the generated key actions and the physically grounded executions.  
    
    \noindent \textbf{Autoregression.}
    To support long-horizon humanoid-object interaction generation, we adopt an autoregressive sampling strategy. Instead of generating the entire sequence in a single forward pass, we generate shorter segmented sequences with overlapping windows. At each autoregressive step, the model generates a short sequence of key actions conditioned on the previous motion. The last few frames of this generated sequence then serve as the initial condition for the subsequent predictive window. This approach enables the scalable generation of long-horizon interactions while maintaining temporal coherence and continuity across windows. Furthermore, because our model predicts sparse key actions while intermediate frames are interpolated, each prediction effectively covers a longer temporal range, thereby significantly reducing the number of required iterations and improving the real-time performance of the generative model. 

\begin{figure}[H]
    \begin{algorithm}[H]
        \textbf{input}: $\bm{\tau}_n$, $\bm{C}$($\bm {\hat G}$, $\bm S$, $\bm {e}_{\text{text}}$); {\color{olive} // input noise and condition information, including object geometry, masked motions and text embedding} \\
        \textbf{output}: $\bm Z_t$; {\color{olive} // joint target torques for the humanoid}\\
        
        \textbf{Step 1: Humanoid-Object Interaction Generation} \\
        \begin{algorithmic}
        \vspace{-0.7em}
            \FOR{each generation}
                \STATE {\color{olive}// concatenate noise and conditions, pass to diffusion model} 
                \STATE $\bm{z} \gets \text{concat}(\bm{\tau}_n, \bm{C})$ 
                \STATE {\color{olive} // generate human motion, object motion, contact prob.} 
                \STATE $(\bm X, \bm O, \bm H) \gets \text{DiffusionModel}(\bm{z})$ 
                \STATE {\color{olive}// use the current generated latest $\bm n_{\text{over}}$ frames as masked motion input for next generation} 
                \STATE $\text{next }\bm{S} \gets \text{Overframes}(\bm X, \bm O, \bm H, \bm n_{\text{over}})$
            \ENDFOR
        \end{algorithmic}
        
        \textbf{Step 2: Interpolation and Conversion} \\
        {\color{olive}// interpolate key actions to obtain dense motion} \\
        $(\bm X_{\text{dense}}, \bm O_{\text{dense}}, \bm H_{\text{dense}}) \gets \text{Interpolate}(\bm X, \bm O, \bm H)$ \\[2pt]
        {\color{olive}// convert the format of dense human motion} \\
        $(\bm {\hat p}_t,\bm {\hat \theta}_t, \bm {\hat p}^{\text{obj}}_t,\bm {\hat \theta}^{\text{obj}}_t, \bm {\hat c}_t) |_{t=1}^T \gets R2G(\bm X_{\text{dense}}, \bm O_{\text{dense}}, \bm H_{\text{dense}})$ \\
        {\color{olive}// the velocities are calculated from kinematic rules} \\
        $(\bm {\hat v}_t,\bm {\hat \omega}_t, \bm {\hat v}^{\text{obj}}_t,\bm {\hat \omega}^{\text{obj}}_t) |_{t=1}^T \gets \text{Kinematic}(\bm {\hat p}_t,\bm {\hat \theta}_t, \bm {\hat p}^{\text{obj}}_t,\bm {\hat \theta}^{\text{obj}}_t)$ \\
        {\color{olive}// reference state variables} \\
        $\bm {\hat s}^p_t = \{\bm {\hat p}_t,\bm {\hat \theta}_t,\bm {\hat v}_t, \bm {\hat \omega}_t\}$ \\
        $\bm {\hat q}^{\text{obj}}_t = \{\bm {\hat p}^{\text{obj}}_t, \bm {\hat \theta}^{\text{obj}}_t, \bm {\hat v}^{\text{obj}}_t, \bm {\hat \omega}^{\text{obj}}_t\}$ \\[2pt]        

        \textbf{Step 3: Contact-Aware Whole Body Control} \\
        {\color{olive}// prepare simulation state variables} \\
        $\bm s^p_t = \{\bm p_t,\bm \theta_t,\bm v_t, \bm \omega_t\}$ \\
        $\bm q^{\text{obj}}_t = \{\bm p^{\text{obj}}_t, \bm \theta^{\text{obj}}_t, \bm v^{\text{obj}}_t, \bm \omega^{\text{obj}}_t\}$ \\[2pt]        
        $\bm Z_t \gets \bm \pi_{\text{HOI}}\big(\bm {\hat s}_{t+1}^p, \bm {\hat q^{\text{obj}}}_{t+1:t+n}, \bm {\hat c}_{t+1}, \bm s^p_t, \bm q^{\text{obj}}_t,\bm{\hat G}\big)$ \\[2pt]
        Apply $\bm Z_t$ to humanoid in simulation \\

        \caption{SimGen-HOI Inference}
        \label{alg:sim_gen_hoi}
    \end{algorithm}
\end{figure}

\captionsetup{justification=centering}
\begin{table*}[t]
\small
\vspace{-5mm}
\caption{\textbf{Evaluation of the generated interaction} on the FullBodyManipulation dataset~\cite{li2023object}. \\
    Units: $T_*$, MPJPE, $H_{feet}$, FS, $P_{hand}$ in millimeters;}

    \label{tab:single_window_cmp_seen}
    \vspace{1mm}
    \centering 
\footnotesize{
\setlength{\tabcolsep}{1pt}
  \resizebox{\textwidth}{!}{ 
\begin{tabular}{lcccccccccccccccc} 
 \toprule 
 & \multicolumn{3}{c}{Condition Matching} & \multicolumn{2}{c}{Human Motion} & \multicolumn{5}{c}{Interaction} & \multicolumn{4}{c}{GT Difference}  \\
 \cmidrule(lr){2-4}\cmidrule(lr){5-6}\cmidrule(lr){7-11}\cmidrule(lr){12-15}  
 Method     & $T_{s}\downarrow$ & $T_{e}\downarrow$ & $T_{xy}\downarrow$ & $H_{feet}\downarrow$ & FS$\downarrow$  & $C_{prec}\uparrow$ & $C_{rec}\uparrow$ & $C_{F_1}\uparrow$ & $C_{\%}$ & $P_{hand}\downarrow$ & MPJPE$\downarrow$  & $T_{root}\downarrow$  & $T_{obj}\downarrow$ & $O_{obj}\downarrow$   \\
\midrule
Interdiff & 0.00 & 158.84 & 72.72 & \textbf{0.90} & 0.42 & 0.63 & 0.28 & 0.33 & 0.27 & 0.55 & 25.91 & 63.44 & 88.35 &  1.65   \\
MDM & 5.18 & 33.07 & 19.42 & 6.72 & 0.48 & 0.72 & 0.47 & 0.53 & 0.43 & 0.66 & 17.86 & 34.16 & 24.46 & 1.85   \\
\midrule 
Lin-OMOMO & 0.00 & 0.00 & 0.00 & 7.21 & 0.41 & 0.68 & 0.56 & 0.57 & 0.54 & {0.51} & 21.73 & 36.62 & 17.12 & 1.21 \\
Pred-OMOMO  &  2.39 & 8.03 & 4.15 & 7.08 & 0.40 & 0.73 & 0.66 & 0.66 & 0.62 & 0.58 & 18.66 & 28.39 & 16.36 & 1.05  \\
GT-OMOMO  & 0.00 & 0.00 & 0.00 & 7.10 & 0.41 & 0.77 & 0.66 & 0.67 & 0.59 & 0.55 & 15.82 & 24.75 & 0.00 & 0.00   \\ 
\midrule
$\text{CHOIS}_{1000}$ & {1.71} & 6.31 & 2.87 & {4.20} & \textbf{0.35} & 0.80 & 0.64 & 0.67 & 0.54 & 0.59 & 15.30 &  24.43 & 12.53 & 0.99 \\
\rowcolor{gray!20}
$\text{OURS}_{200} $ w/o $w_i$ & 1.85 & 4.58 & 3.67 & 6.22 & 0.68 & 0.87 & 0.41 & 0.50 & 0.37 & \textbf{0.50} & 12.0 &  15.35 & 9.25 & 0.93 \\
\rowcolor{gray!20}
$\text{OURS}_{200}$ & 1.65 & 2.87 & 1.95 & 6.92 & 0.44 & \textbf{0.89} & \textbf{0.74} & \textbf{0.77} & 0.70 & 0.54 & 7.67 &  9.80 & 4.39 & \textbf{0.38} \\
\rowcolor{gray!20}
$\text{OURS}_{1000}$ & \textbf{1.54} & \textbf{2.72} & \textbf{1.83} & 6.86 & {0.44} & 0.88 & {0.73} & {0.76} & 0.70 & 0.54 & \textbf{7.56} &  \textbf{9.71} & \textbf{4.32} & {0.39} \\
\bottomrule
\end{tabular}
}
}
\vspace{-2mm}
\end{table*}

\section{Experiments}
We first describe training and evaluation datasets, performance metrics, and baseline comparisons to validate our key action-based generation strategy. We then show the policy’s tracking capability metrics and the physical metrics of the action, and finally verify the impact of fine-tuning on the success rate of the policy’s tracking generation results.

\subsection{Datasets}
    \noindent\textbf{FullBodyManipulation Dataset.}
    Following Chois~\cite {chois2024}, we use the FullBodyManipulation dataset~\cite{li2023object}, which contains 10 hours of high-quality paired human-object interaction data involving 15 different objects. The dataset is used for both training and evaluation. Following the same data split as OMOMO~\cite{li2023object}, we train on data from 15 subjects and reserve two subjects for testing. For contact-aware policy training, we utilize a robot modeled with the SMPL-X body model~\cite{smplx}, which includes articulated hand joints. 

\subsection{Evaluation Metrics.}
    \noindent\textbf{Generation Metrics.} We evaluate our method using four categories of metrics. \textbf{Condition matching} measures how well the generated object trajectory aligns with input object waypoints via Euclidean distances at the start ($T_s$), end ($T_e$), and intermediate positions ($T_{xy}$). \textbf{Human motion quality} is assessed using standard metrics: foot sliding (FS) and foot height ($H_{feet}$) indicate motion smoothness and contact plausibility~\cite{he2022nemf}. \textbf{Interaction quality} is evaluated by contact precision ($C_{prec}$), recall ($C_{rec}$), F1 score ($C_{F_1}$), and contact percentage ($C_{\%}$), along with a Signed Distance Field-based hand penetration score ($P_{hand}$) that quantifies physical plausibility~\cite{li2023object}. Finally, \textbf{Deviations from Ground truth}, consisting of  MPJPE, root translation ($T_{root}$), object position errors ($T_{obj}$), as well as orientation errors ($O_{root}$, $O_{obj}$) using Frobenius norms, are reported. Together, these metrics comprehensively evaluate motion accuracy, realism, and interaction fidelity.
    
    \noindent\textbf{Motion Tracking Metrics.} To enable the comparisons between different motion tracking policies, we adopt the evaluation metrics defined in OmniGrasp~\cite{omnigrasp2024}, including humanoid position error ($E_\text{pos}$, mm), humanoid rotation error ($E_\text{rot}$, rad), and objects position and rotation error, $E_\text{pos}^{obj}$ and $E_\text{rot}^{obj}$. To evaluate the interaction performance, we report the contact success rate ($\text{Succ}_\text{cont}$), which considers whether the object is contacted by the correct part of the humanoid's body. We also include tracking duration to calculate the entire process without triggering the termination condition of the object being too far away. Furthermore, a success rate ($\text{Succ}_\text{tgt}$) to evaluate the object has reached its final position within a distance of 0.5 meters.
    
\begin{table}[tp]
    \centering
    \caption{ Evaluation of motions tracked by our control policy on the FullBodyManipulation dataset.} 
    \label{tab:RL_foot_sliding}
    \vspace{0mm}
    \centering 
    \footnotesize
    \setlength{\tabcolsep}{14pt}
    
    \begin{tabular}{lccc} 
         \toprule 
        & \multicolumn{3}{c}{Physical Consistency}\\
         \cmidrule(lr){2-4}
         Method & $H_{feet}\downarrow$ & FS$\downarrow$ & $P_{hand}\downarrow$ \\
        \midrule
        OMOMO GT & 3.4 & 0.28 & 0.56\\
        CHOIS & 4.20 & 0.35 & 0.59\\
        \rowcolor{gray!20}
        $\text{OURS}_{200}$ & 6.92 & 0.43 & 0.54 \\
        \rowcolor{gray!20}
        SimGen-HOI &  \textbf{1.77} & \textbf{0.13} & \textbf{0.00} \\
        \bottomrule
    \end{tabular}
    
    \vspace{-2mm}
\end{table}

\captionsetup{justification=centering}
\begin{table*}[t!]
    \small
    \vspace{-7mm}
    \caption{Different combination methods\\
            Units: $E_{\text{pos}}^{obj}$ mm, $E_{\text{rot}}^{obj}$ rad, $\text{E}_{\text{acc}}^{obj}$ $\text{mm} / \text{s}^2$, $\text{E}_{\text{vel}}^{obj}$ $\text{mm} / \text{s}$;} 
    \label{tab:diff_combine}
    \centering 
    \footnotesize
    \setlength{\tabcolsep}{13pt} 
    
    \begin{tabular}{lcccccccc} 
    \toprule 
    & \multicolumn{7}{c}{OMOMO (7 objects in GT)} \\
    \cmidrule(lr){2-8}
    Method & $\text{Succ}_\text{cont} \uparrow$ & $\text{Succ}_\text{tgt} \uparrow$ & $TTR \uparrow$ & $E_{\text{pos}}^{obj} \downarrow$ & $E_{\text{rot}}^{obj} \downarrow$ & $\text{E}_{\text{acc}}^{obj} \downarrow$ & $\text{E}_{\text{vel}}^{obj} \downarrow$\\ 
    \midrule
    Chois + PHC-X & 0/7 & 0/7 & 29.40\% & 748.03 & 1.88 & 3.30 & 15.61 \\
    OmniGrasp on OMOMO & 6/7 & \textbf{7/7} & \textbf{95.00}\% & \textbf{40.84} & \textbf{0.13} & 3.21 & 8.91 \\
    \rowcolor{gray!20}
    SimGen-HOI & \textbf{7/7} & \textbf{7/7} & 92.10\% & 64.92 & 0.72 & \textbf{3.05} & \textbf{5.80} \\
    \bottomrule
    \end{tabular}
    \vspace{-2mm}
\end{table*}



\begin{table}[tp]
    \centering
    \caption{Ablation Study} 
    \label{tab:ablation}
    \vspace{0mm}
    \centering 
    \footnotesize
    \setlength{\tabcolsep}{8pt}
        \begin{tabular}{lcccccccc} 
        \toprule 
        & \multicolumn{4}{c}{OMOMO (smallbox in Generation)} \\
        \cmidrule(lr){2-5}
        Method & $\text{Succ} \uparrow$ & $E_\text{pos}^{obj} \downarrow$ & $E_\text{rot}^{obj} \downarrow$ & $\text{E}_{\text{pos}} \downarrow$ \\ 
        \midrule
        w.o. key action & 34.72\% & 128.00 & 0.39 & 111.89\\
        w.o. contact & 31.94\% & 115.93 & 0.26 & \textbf{73.70}\\
        w.o. finetune & 37.50\% & 119.53 & 0.31 & 108.41\\
        \rowcolor{gray!20}
        Full & \textbf{41.67\%} & \textbf{114.23} & \textbf{0.25} & 102.92\\
        \bottomrule
        \end{tabular}
    \vspace{-2mm}
\end{table}

\subsection{Results.}
    \noindent \textbf{Baselines.}
    To evaluate the effectiveness of our humanoid-object motion generation, we compare against several adapted baselines from related tasks followed by~\citep{chois2024}. InterDiff~\cite{xu2023interdiff}, MDM~\cite{tevet2022human}, and OMOMO~\cite{li2023object} are modified to fit our mission settings. Since our method predicts only a sparse set of key actions, we recover the full motion sequence by interpolating these key actions at the known data indices for fair comparison. As shown in Table~\ref{tab:single_window_cmp_seen}, our method (OURS) consistently outperforms baselines across multiple evaluation aspects about matching accuracy, interaction quality, and ground truth difference. The subscripts on OURS indicate the number of denoising steps in the diffusion model. Continually increasing the number of steps yields diminishing improvements, suggesting that model performance saturates beyond a certain step size. The variant w/o $\bm w_i$ (in Eq.\eqref{eq:reconstruction_error}) refers to the model without weights during key actions extraction. This results in a notable performance drop across several metrics. Interestingly, hand penetration is lower as the model fails to capture fine-grained contact, causing hands to stay farther from objects and lower mesh collisions. Meanwhile, due to its inherent limitations, the interpolation-based method performs worse on physics-related metrics, particularly foot motion. However, when the motions are tracked into the physics simulation later, the physics-related metrics are improved significantly.

    To validate the benefits of incorporating physics simulation into humanoid-object motion generation, we compare physical plausibility metrics of the motion. Specifically, we evaluate foot height consistency ($H_{\text{feet}}$) and foot sliding (FS), as reported in Table~\ref{tab:RL_foot_sliding}. The results demonstrate that incorporating simulation significantly improves the physical realism of humanoid motion, notably reducing foot sliding and hand-object penetration artifacts.

\subsection{Contact-Aware Interaction Tracking.}
    To compare the tracking ability of different methods in physics simulation, we evaluate three methods in 7 reference motions as OmniGrasp~\cite{omnigrasp2024}, which are selected manually. Since the OMOMO~\cite{li2023object} dataset does not provide detailed hand motions, which are required for OmniGrasp training. We employ the GRIP~\cite{taheri2024grip} method to refine the OMOMO dataset and generate detailed hand grasping motions as shown in Figure~\ref{fig:qualitative_res}. As the CHOIS~\cite{chois2024} model also lacks detailed hand motions, we refine it in the same way as the OMOMO dataset. Then apply the SMPL-X tracker PHC-X~\cite{Luo2023PerpetualHC} to track the refined motions. For our methods, we convert the dataset into a format compatible with our policy and assign default joint angles to the missing hand articulations, which do not need a detailed grasp pose. The tracking results in Table~\ref{tab:diff_combine} show that directly tracking generated motion often fails to achieve reliable object interaction due to its high sensitivity to motion accuracy. OmniGrasp fine-tuned on OMOMO demonstrates strong trajectory-following capabilities for grasping because it only tracks object trajectories and the grasping motion is explored by RL rather than learned from reference motions. In addition, OmniGrasp relies heavily on pre-grasped poses when grasping large objects, The vulnerability to inaccuracies in the poses between objects and hands makes it difficult to achieve improvement through RL.

\subsection{Ablation Study.}
    To verify the contribution of the proposed method to the HOI task, we conducted an ablation experiment on the main modules we proposed. The experiments are conducted on all the small boxes sequences (in total 144 sequences) from the test set of the OMOMO~\citep{li2023object} dataset. 
    
    \noindent \textbf{Key Actions Design.}
    We evaluate the effect of the key actions extraction and interpolation strategy by removing it and predicting dense motion sequences directly instead. The policy tracking results are shown in Table~\ref{tab:ablation}. The results suggest lower overall tracking success and reduced motion fidelity. Without key actions, the model's output requires more inference steps and produces less structured motion, making policy tracking more difficult and reducing both humanoid and object tracking accuracy.
    
    \noindent \textbf{Contact Guidance.}
    To assess the role of contact cues, we remove contact guidance from the policy input. This results in a clear drop in object tracking precision and a lower rate of successful interactions. The lack of contact signals hinders the strategy's ability to grasp and release events in time, making it difficult for the agent to learn through RL. Without the contact guidance, the humanoid prioritizes tracking the reference humanoid motion, which leads to a lower humanoid position error.

    \noindent \textbf{Fine-tuning.}
    We disable the alternating fine-tuning as described in the Methods section to test its influence. As reported in Table~\ref{tab:ablation}, the tracking success rate decreases in the experiment without fine-tuning. 
\section{Conclusion}
    \label{sec:conclusion}
    In conclusion, our work tackles the challenge of generating physically realistic humanoid-object interactions. By employing the concept of un-uniform key actions, the generation model is changed to predict a sequence of crucial actions. It significantly improves the performance of generation and naturally supports long-horizon action generation. We further propose an effective contact-aware policy for tracking the generated motions in physically realistic environments, even in long-horizon tasks, which provides strong support for a broad range of object manipulation capabilities. An alternating fine-tuning paradigm is introduced to enhance the tracking success rate, enabling our autoregressive framework to successfully execute complex, long-horizon interaction tasks specified by the user in a 3D environment.

\bigskip
\bibliography{a2026}

\begin{thebibliography}{48}
\providecommand{\natexlab}[1]{#1}

\bibitem[{Bae et~al.(2023)Bae, Won, Lim, Min, and Kim}]{bae2023pmp}
Bae, J.; Won, J.; Lim, D.; Min, C.-H.; and Kim, Y.~M. 2023.
\newblock Pmp: Learning to physically interact with environments using part-wise motion priors.
\newblock In \emph{ACM SIGGRAPH 2023 Conference Proceedings}, 1--10.

\bibitem[{Braun et~al.(2024)Braun, Christen, Kocabas, Aksan, and Hilliges}]{braun2024physically}
Braun, J.; Christen, S.; Kocabas, M.; Aksan, E.; and Hilliges, O. 2024.
\newblock Physically plausible full-body hand-object interaction synthesis.
\newblock In \emph{2024 International Conference on 3D Vision (3DV)}, 464--473. IEEE.

\bibitem[{Cen et~al.(2024)Cen, Pi, Peng, Shen, Yang, Zhu, Bao, and Zhou}]{cen2024generating}
Cen, Z.; Pi, H.; Peng, S.; Shen, Z.; Yang, M.; Zhu, S.; Bao, H.; and Zhou, X. 2024.
\newblock Generating human motion in 3D scenes from text descriptions.
\newblock In \emph{Proceedings of the IEEE/CVF Conference on Computer Vision and Pattern Recognition}, 1855--1866.

\bibitem[{Chen et~al.(2024)Chen, Shi, Huang, Tan, Komura, and Chen}]{chen2024taming}
Chen, R.; Shi, M.; Huang, S.; Tan, P.; Komura, T.; and Chen, X. 2024.
\newblock Taming diffusion probabilistic models for character control.
\newblock In \emph{ACM SIGGRAPH 2024 Conference Papers}, 1--10.

\bibitem[{Chentanez et~al.(2018)Chentanez, M{\"u}ller, Macklin, Makoviychuk, and Jeschke}]{chentanez2018physics}
Chentanez, N.; M{\"u}ller, M.; Macklin, M.; Makoviychuk, V.; and Jeschke, S. 2018.
\newblock Physics-based motion capture imitation with deep reinforcement learning.
\newblock In \emph{Proceedings of the 11th ACM SIGGRAPH Conference on Motion, Interaction and Games}, 1--10.

\bibitem[{Dabral et~al.(2023)Dabral, Mughal, Golyanik, and Theobalt}]{dabral2023mofusion}
Dabral, R.; Mughal, M.~H.; Golyanik, V.; and Theobalt, C. 2023.
\newblock Mofusion: A framework for denoising-diffusion-based motion synthesis.
\newblock In \emph{Proceedings of the IEEE/CVF conference on computer vision and pattern recognition}, 9760--9770.

\bibitem[{Guo et~al.(2022)Guo, Zou, Zuo, Wang, Ji, Li, and Cheng}]{Guo_2022_CVPR}
Guo, C.; Zou, S.; Zuo, X.; Wang, S.; Ji, W.; Li, X.; and Cheng, L. 2022.
\newblock Generating Diverse and Natural {3D} Human Motions From Text.
\newblock In \emph{CVPR}.

\bibitem[{He et~al.(2022)He, Saito, Zachary, Rushmeier, and Zhou}]{he2022nemf}
He, C.; Saito, J.; Zachary, J.; Rushmeier, H.; and Zhou, Y. 2022.
\newblock Nemf: Neural motion fields for kinematic animation.
\newblock \emph{NeurIPS}.

\bibitem[{He et~al.(2024)He, Luo, He, Xiao, Zhang, Zhang, Kitani, Liu, and Shi}]{he2024omnih2o}
He, T.; Luo, Z.; He, X.; Xiao, W.; Zhang, C.; Zhang, W.; Kitani, K.; Liu, C.; and Shi, G. 2024.
\newblock Omnih2o: Universal and dexterous human-to-humanoid whole-body teleoperation and learning.
\newblock \emph{arXiv preprint arXiv:2406.08858}.

\bibitem[{Ho, Jain, and Abbeel(2020)}]{ddpm}
Ho, J.; Jain, A.; and Abbeel, P. 2020.
\newblock Denoising diffusion probabilistic models.
\newblock \emph{Advances in neural information processing systems}, 33: 6840--6851.

\bibitem[{Huang et~al.(2023)Huang, Wang, Li, Jia, Liu, Zhu, Liang, and Zhu}]{huang2023diffusion}
Huang, S.; Wang, Z.; Li, P.; Jia, B.; Liu, T.; Zhu, Y.; Liang, W.; and Zhu, S.-C. 2023.
\newblock Diffusion-based generation, optimization, and planning in 3d scenes.
\newblock In \emph{Proceedings of the IEEE/CVF Conference on Computer Vision and Pattern Recognition}, 16750--16761.

\bibitem[{Jiang et~al.(2024{\natexlab{a}})Jiang, He, Wang, Li, Chen, Huang, and Zhu}]{lingo2024}
Jiang, N.; He, Z.; Wang, Z.; Li, H.; Chen, Y.; Huang, S.; and Zhu, Y. 2024{\natexlab{a}}.
\newblock Autonomous character-scene interaction synthesis from text instruction.
\newblock In \emph{SIGGRAPH Asia 2024 Conference Papers}, 1--11.

\bibitem[{Jiang et~al.(2024{\natexlab{b}})Jiang, Zhang, Li, Ma, Wang, Chen, Liu, Zhu, and Huang}]{truman2024}
Jiang, N.; Zhang, Z.; Li, H.; Ma, X.; Wang, Z.; Chen, Y.; Liu, T.; Zhu, Y.; and Huang, S. 2024{\natexlab{b}}.
\newblock Scaling up dynamic human-scene interaction modeling.
\newblock In \emph{Proceedings of the IEEE/CVF Conference on Computer Vision and Pattern Recognition}, 1737--1747.

\bibitem[{Kulkarni et~al.(2024)Kulkarni, Rempe, Genova, Kundu, Johnson, Fouhey, and Guibas}]{kulkarni2024nifty}
Kulkarni, N.; Rempe, D.; Genova, K.; Kundu, A.; Johnson, J.; Fouhey, D.; and Guibas, L. 2024.
\newblock Nifty: Neural object interaction fields for guided human motion synthesis.
\newblock In \emph{Proceedings of the IEEE/CVF Conference on Computer Vision and Pattern Recognition}, 947--957.

\bibitem[{Li et~al.(2024{\natexlab{a}})Li, Clegg, Mottaghi, Wu, Puig, and Liu}]{chois2024}
Li, J.; Clegg, A.; Mottaghi, R.; Wu, J.; Puig, X.; and Liu, C.~K. 2024{\natexlab{a}}.
\newblock Controllable human-object interaction synthesis.
\newblock In \emph{ECCV}.

\bibitem[{Li et~al.(2024{\natexlab{b}})Li, Clegg, Mottaghi, Wu, Puig, and Liu}]{li2024controllable}
Li, J.; Clegg, A.; Mottaghi, R.; Wu, J.; Puig, X.; and Liu, C.~K. 2024{\natexlab{b}}.
\newblock Controllable human-object interaction synthesis.
\newblock In \emph{European Conference on Computer Vision}, 54--72. Springer.

\bibitem[{Li, Wu, and Liu(2023)}]{li2023object}
Li, J.; Wu, J.; and Liu, C.~K. 2023.
\newblock Object motion guided human motion synthesis.
\newblock \emph{ACM Transactions on Graphics (TOG)}, 42(6): 1--11.

\bibitem[{Liu and Hodgins(2018)}]{liu2018learning}
Liu, L.; and Hodgins, J. 2018.
\newblock Learning basketball dribbling skills using trajectory optimization and deep reinforcement learning.
\newblock \emph{Acm transactions on graphics (tog)}, 37(4): 1--14.

\bibitem[{Luo et~al.(2024)Luo, Cao, Christen, Winkler, Kitani, and Xu}]{omnigrasp2024}
Luo, Z.; Cao, J.; Christen, S.; Winkler, A.; Kitani, K.; and Xu, W. 2024.
\newblock Omnigrasp: Grasping diverse objects with simulated humanoids.
\newblock \emph{Advances in Neural Information Processing Systems}, 37: 2161--2184.

\bibitem[{Luo et~al.(2023{\natexlab{a}})Luo, Cao, Merel, Winkler, Huang, Kitani, and Xu}]{luo2023universal}
Luo, Z.; Cao, J.; Merel, J.; Winkler, A.; Huang, J.; Kitani, K.; and Xu, W. 2023{\natexlab{a}}.
\newblock Universal humanoid motion representations for physics-based control.
\newblock \emph{arXiv preprint arXiv:2310.04582}.

\bibitem[{Luo et~al.(2023{\natexlab{b}})Luo, Cao, Winkler, Kitani, and Xu}]{Luo2023PerpetualHC}
Luo, Z.; Cao, J.; Winkler, A.~W.; Kitani, K.; and Xu, W. 2023{\natexlab{b}}.
\newblock Perpetual Humanoid Control for Real-time Simulated Avatars.
\newblock In \emph{International Conference on Computer Vision (ICCV)}.

\bibitem[{Luo, Yuan, and Kitani(2022)}]{luo2022universal}
Luo, Z.; Yuan, Y.; and Kitani, K.~M. 2022.
\newblock From universal humanoid control to automatic physically valid character creation.
\newblock \emph{arXiv preprint arXiv:2206.09286}.

\bibitem[{Mahmood et~al.(2019)Mahmood, Ghorbani, Troje, Pons-Moll, and Black}]{AMASS}
Mahmood, N.; Ghorbani, N.; Troje, N.~F.; Pons-Moll, G.; and Black, M.~J. 2019.
\newblock {AMASS}: Archive of motion capture as surface shapes.
\newblock In \emph{ICCV}.

\bibitem[{Merel et~al.(2020)Merel, Tunyasuvunakool, Ahuja, Tassa, Hasenclever, Pham, Erez, Wayne, and Heess}]{merel2020catch}
Merel, J.; Tunyasuvunakool, S.; Ahuja, A.; Tassa, Y.; Hasenclever, L.; Pham, V.; Erez, T.; Wayne, G.; and Heess, N. 2020.
\newblock Catch \& carry: reusable neural controllers for vision-guided whole-body tasks.
\newblock \emph{ACM Transactions on Graphics (TOG)}, 39(4): 39--1.

\bibitem[{Pavlakos et~al.(2019)Pavlakos, Choutas, Ghorbani, Bolkart, Osman, Tzionas, and Black}]{smplx}
Pavlakos, G.; Choutas, V.; Ghorbani, N.; Bolkart, T.; Osman, A. A.~A.; Tzionas, D.; and Black, M.~J. 2019.
\newblock Expressive Body Capture: 3{D} Hands, Face, and Body From a Single Image.
\newblock In \emph{CVPR}.

\bibitem[{Peebles and Xie(2022)}]{DiT}
Peebles, W.; and Xie, S. 2022.
\newblock Scalable Diffusion Models with Transformers.
\newblock \emph{International Conference on Computer Vision}.

\bibitem[{Peng et~al.(2023)Peng, Xie, Wu, Jampani, Sun, and Jiang}]{peng2023hoi}
Peng, X.; Xie, Y.; Wu, Z.; Jampani, V.; Sun, D.; and Jiang, H. 2023.
\newblock Hoi-diff: Text-driven synthesis of 3d human-object interactions using diffusion models.
\newblock \emph{arXiv preprint arXiv:2312.06553}.

\bibitem[{Peng et~al.(2018)Peng, Abbeel, Levine, and Van~de Panne}]{peng2018deepmimic}
Peng, X.~B.; Abbeel, P.; Levine, S.; and Van~de Panne, M. 2018.
\newblock Deepmimic: Example-guided deep reinforcement learning of physics-based character skills.
\newblock \emph{ACM Transactions On Graphics (TOG)}, 37(4): 1--14.

\bibitem[{Prokudin, Lassner, and Romero(2019)}]{bps}
Prokudin, S.; Lassner, C.; and Romero, J. 2019.
\newblock Efficient learning on point clouds with basis point sets.
\newblock In \emph{ICCV}.

\bibitem[{Punnakkal et~al.(2021)Punnakkal, Chandrasekaran, Athanasiou, Quiros-Ramirez, and Black}]{BABEL}
Punnakkal, A.~R.; Chandrasekaran, A.; Athanasiou, N.; Quiros-Ramirez, A.; and Black, M.~J. 2021.
\newblock {BABEL}: Bodies, Action and Behavior with English Labels.
\newblock In \emph{CVPR}.

\bibitem[{Raab et~al.(2024)Raab, Gat, Sala, Tevet, Shalev-Arkushin, Fried, Bermano, and Cohen-Or}]{raab2024monkey}
Raab, S.; Gat, I.; Sala, N.; Tevet, G.; Shalev-Arkushin, R.; Fried, O.; Bermano, A.~H.; and Cohen-Or, D. 2024.
\newblock Monkey see, monkey do: Harnessing self-attention in motion diffusion for zero-shot motion transfer.
\newblock In \emph{SIGGRAPH Asia 2024 Conference Papers}, 1--13.

\bibitem[{Raab et~al.(2023)Raab, Leibovitch, Tevet, Arar, Bermano, and Cohen-Or}]{raab2023single}
Raab, S.; Leibovitch, I.; Tevet, G.; Arar, M.; Bermano, A.~H.; and Cohen-Or, D. 2023.
\newblock Single motion diffusion.
\newblock \emph{arXiv preprint arXiv:2302.05905}.

\bibitem[{Radford et~al.(2021)Radford, Kim, Hallacy, Ramesh, Goh, Agarwal, Sastry, Askell, Mishkin, Clark et~al.}]{clip}
Radford, A.; Kim, J.~W.; Hallacy, C.; Ramesh, A.; Goh, G.; Agarwal, S.; Sastry, G.; Askell, A.; Mishkin, P.; Clark, J.; et~al. 2021.
\newblock Learning transferable visual models from natural language supervision.
\newblock In \emph{ICML}.

\bibitem[{Romero, Tzionas, and Black(2022)}]{romero2022embodied}
Romero, J.; Tzionas, D.; and Black, M.~J. 2022.
\newblock Embodied hands: Modeling and capturing hands and bodies together.
\newblock \emph{arXiv preprint arXiv:2201.02610}.

\bibitem[{Shi et~al.(2023)Shi, Sharma, Zhao, and Finn}]{shi2023waypoint}
Shi, L.~X.; Sharma, A.; Zhao, T.~Z.; and Finn, C. 2023.
\newblock Waypoint-based imitation learning for robotic manipulation.
\newblock \emph{arXiv preprint arXiv:2307.14326}.

\bibitem[{Shi et~al.(2024)Shi, Wang, Jiang, Lin, Dai, and Peng}]{shi2024interactive}
Shi, Y.; Wang, J.; Jiang, X.; Lin, B.; Dai, B.; and Peng, X.~B. 2024.
\newblock Interactive character control with auto-regressive motion diffusion models.
\newblock \emph{ACM Transactions on Graphics (TOG)}, 43(4): 1--14.

\bibitem[{Taheri et~al.(2024)Taheri, Zhou, Tzionas, Zhou, Ceylan, Pirk, and Black}]{taheri2024grip}
Taheri, O.; Zhou, Y.; Tzionas, D.; Zhou, Y.; Ceylan, D.; Pirk, S.; and Black, M.~J. 2024.
\newblock Grip: Generating interaction poses using spatial cues and latent consistency.
\newblock In \emph{2024 International Conference on 3D Vision (3DV)}, 933--943. IEEE.

\bibitem[{Tevet et~al.(2024)Tevet, Raab, Cohan, Reda, Luo, Peng, Bermano, and van~de Panne}]{Closd2024}
Tevet, G.; Raab, S.; Cohan, S.; Reda, D.; Luo, Z.; Peng, X.~B.; Bermano, A.~H.; and van~de Panne, M. 2024.
\newblock CLoSD: Closing the Loop between Simulation and Diffusion for multi-task character control.
\newblock \emph{arXiv preprint arXiv:2410.03441}.

\bibitem[{Tevet et~al.(2022)Tevet, Raab, Gordon, Shafir, Cohen-Or, and Bermano}]{tevet2022human}
Tevet, G.; Raab, S.; Gordon, B.; Shafir, Y.; Cohen-Or, D.; and Bermano, A.~H. 2022.
\newblock Human motion diffusion model.
\newblock \emph{arXiv preprint arXiv:2209.14916}.

\bibitem[{Vaswani et~al.(2017)Vaswani, Shazeer, Parmar, Uszkoreit, Jones, Gomez, Kaiser, and Polosukhin}]{transformer}
Vaswani, A.; Shazeer, N.; Parmar, N.; Uszkoreit, J.; Jones, L.; Gomez, A.~N.; Kaiser, {\L}.; and Polosukhin, I. 2017.
\newblock Attention is all you need.
\newblock In \emph{Advances in Neural Information Processing Systems (NIPS)}.

\bibitem[{Wang et~al.(2023)Wang, Lin, Zeng, Luo, Zhang, and Zhang}]{wang2023physhoi}
Wang, Y.; Lin, J.; Zeng, A.; Luo, Z.; Zhang, J.; and Zhang, L. 2023.
\newblock Physhoi: Physics-based imitation of dynamic human-object interaction.
\newblock \emph{arXiv preprint arXiv:2312.04393}.

\bibitem[{Wu et~al.(2024{\natexlab{a}})Wu, Shi, Huang, Yu, Xu, and Wang}]{wu2024thor}
Wu, Q.; Shi, Y.; Huang, X.; Yu, J.; Xu, L.; and Wang, J. 2024{\natexlab{a}}.
\newblock Thor: Text to human-object interaction diffusion via relation intervention.
\newblock \emph{arXiv preprint arXiv:2403.11208}.

\bibitem[{Wu et~al.(2024{\natexlab{b}})Wu, Li, Xu, and Liu}]{hoifhli2024}
Wu, Z.; Li, J.; Xu, P.; and Liu, C.~K. 2024{\natexlab{b}}.
\newblock Human-object interaction from human-level instructions.
\newblock \emph{arXiv preprint arXiv:2406.17840}.

\bibitem[{Xu et~al.(2023)Xu, Li, Wang, and Gui}]{xu2023interdiff}
Xu, S.; Li, Z.; Wang, Y.-X.; and Gui, L.-Y. 2023.
\newblock Interdiff: Generating 3d human-object interactions with physics-informed diffusion.
\newblock In \emph{Proceedings of the IEEE/CVF International Conference on Computer Vision}, 14928--14940.

\bibitem[{Yi et~al.(2024)Yi, Thies, Black, Peng, and Rempe}]{yi2024generating}
Yi, H.; Thies, J.; Black, M.~J.; Peng, X.~B.; and Rempe, D. 2024.
\newblock Generating human interaction motions in scenes with text control.
\newblock In \emph{European Conference on Computer Vision}, 246--263. Springer.

\bibitem[{Zhang et~al.(2024)Zhang, Cai, Pan, Hong, Guo, Yang, and Liu}]{zhang2024motiondiffuse}
Zhang, M.; Cai, Z.; Pan, L.; Hong, F.; Guo, X.; Yang, L.; and Liu, Z. 2024.
\newblock Motiondiffuse: Text-driven human motion generation with diffusion model.
\newblock \emph{IEEE transactions on pattern analysis and machine intelligence}, 46(6): 4115--4128.

\bibitem[{Zhang et~al.(2022)Zhang, Bhatnagar, Starke, Guzov, and Pons-Moll}]{zhang2022couch}
Zhang, X.; Bhatnagar, B.~L.; Starke, S.; Guzov, V.; and Pons-Moll, G. 2022.
\newblock Couch: Towards controllable human-chair interactions.
\newblock In \emph{European Conference on Computer Vision}, 518--535. Springer.

\bibitem[{Zhou et~al.(2019)Zhou, Barnes, Lu, Yang, and Li}]{zhou2019continuity}
Zhou, Y.; Barnes, C.; Lu, J.; Yang, J.; and Li, H. 2019.
\newblock On the continuity of rotation representations in neural networks.
\newblock In \emph{CVPR}.

\end{thebibliography}

\clearpage


\end{document}